\documentclass[letterpaper, 10 pt, conference]{ieeeconf}  

\usepackage{booktabs}
\usepackage{graphicx} 
\usepackage{url}
\usepackage{hyperref}

\usepackage{marvosym}  

\usepackage{amsmath}

\IEEEoverridecommandlockouts                              

\title{\LARGE \bf
Iterative Residual Cross-Attention Mechanism: An Integrated Approach for Audio-Visual Navigation Tasks
}

\author{
Hailong Zhang$^1$, Yinfeng Yu$^{1}$$^{,\mbox{\Letter}}$, Liejun Wang$^1$, Fuchun Sun$^2$, and Wendong Zheng$^3$%
\thanks{$^{\mbox{\Letter}}$\small Yinfeng Yu is the corresponding author (e-mail: yuyinfeng@xju.edu.cn).}
\\
$^1$Xinjiang Multimodal Intelligent Processing and Information Security Engineering Technology Research Center,\\
School of Computer Science and Technology, Xinjiang University, Urumqi 830017, China
\\
$^2$Department of Computer Science and Technology, Tsinghua University, Beijing 100091, China
\\
$^3$School of Electrical Engineering and Automation, Tianjin University of Technology, Tianjin 300382, China%
}

\begin{document}

\maketitle

\thispagestyle{empty}
\pagestyle{empty}

\begin{abstract}

Audio-visual navigation represents a significant area of research in which intelligent agents utilize egocentric visual and auditory perceptions to identify audio targets. Conventional navigation methodologies typically adopt a staged modular design, which involves first executing feature fusion, then utilizing Gated Recurrent Unit (GRU) modules for sequence modeling, and finally making decisions through reinforcement learning. While this modular approach has demonstrated effectiveness, it may also lead to redundant information processing and inconsistencies in information transmission between the various modules during the feature fusion and GRU sequence modeling phases. This paper presents IRCAM-AVN (Iterative Residual Cross-Attention Mechanism for Audiovisual Navigation), an end-to-end framework that integrates multimodal information fusion and sequence modeling within a unified IRCAM module, thereby replacing the traditional separate components for fusion and GRU. This innovative mechanism employs a multi-level residual design that concatenates initial multimodal sequences with processed information sequences. This methodological shift progressively optimizes the feature extraction process while reducing model bias and enhancing the model's stability and generalization capabilities. Empirical results indicate that intelligent agents employing the iterative residual cross-attention mechanism exhibit superior navigation performance.

\end{abstract}

\section{INTRODUCTION}

\begin{figure}[!h] 
    \centering 
    \includegraphics[width=0.5\textwidth]{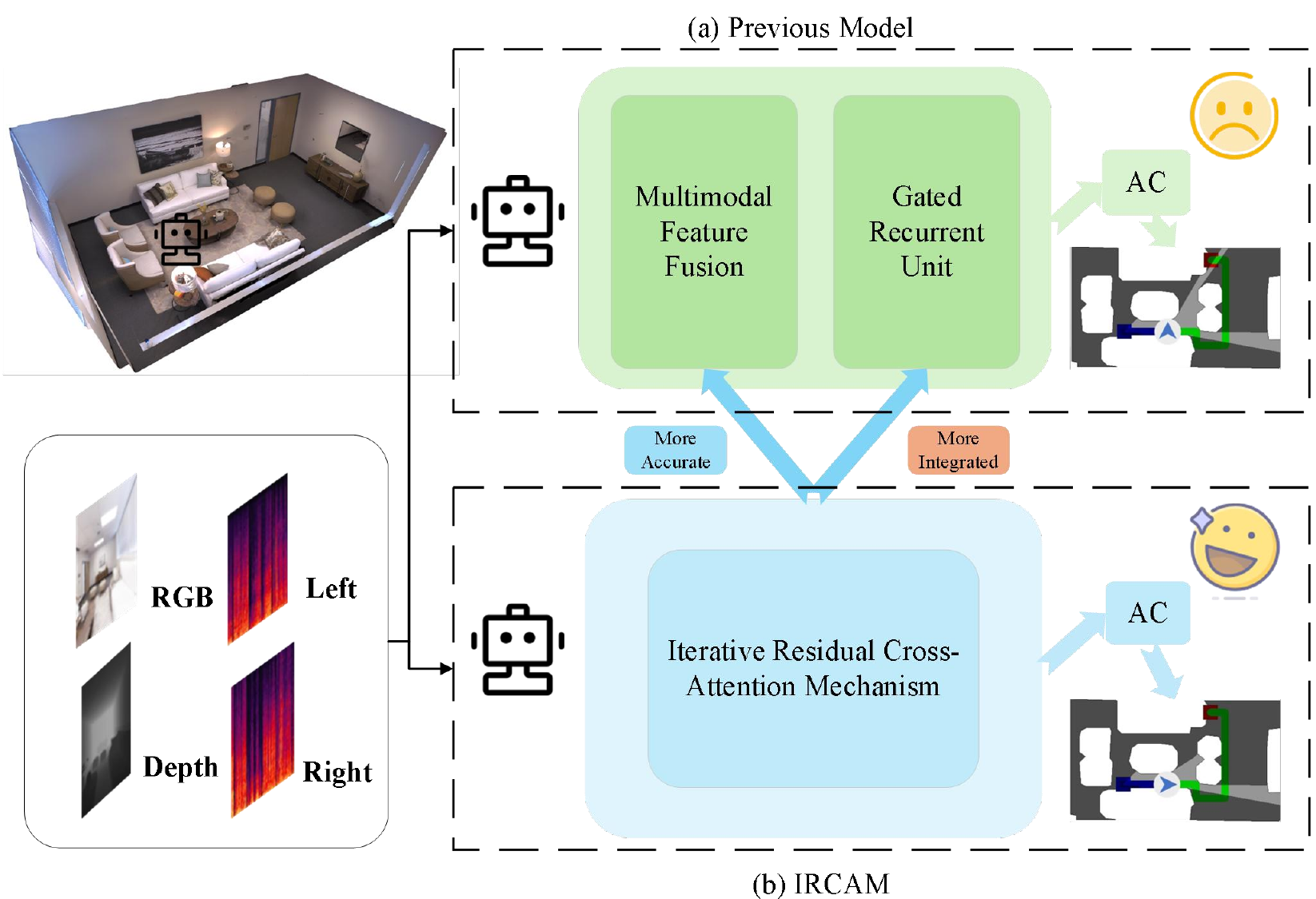}
    \caption{
    A sketched comparison between the previous mainstream method (a) and the proposed IRCAM (b).
    }
    \label{fig1}
\end{figure}

Audiovisual navigation is a critical component in various real-world applications that demand efficient target acquisition. However, intelligent agents often encounter challenges in locating audio cues successfully. Given the unique characteristics of audiovisual navigation tasks, both the success rate and the timeliness of reaching these targets are particularly paramount in numerous scenarios. For instance, in emergencies where an individual calls for assistance, it is essential that the agent rapidly and accurately identifies and reaches the location. Similarly, when a gas alarm activates, the agent must promptly discern and address the source to deactivate the gas valve. Furthermore, if the sound of water running indicates an open faucet, the agent's swift arrival at the water source is crucial. These scenarios underscore the significance and urgency associated with audiovisual navigation, thereby establishing it as a vital area for research.

Previous studies have investigated various methodologies to enhance agents' capabilities in reaching audio targets. These studies have employed feature fusion modules and GRU~\cite{cho2014learning} (Gated Recurrent Unit) to process multimodal information, resulting in relatively high success rates in audiovisual navigation tasks. Despite these advancements, there is still potential for improvement in navigation efficiency and model integration. Challenges such as redundancy in information processing and inconsistencies during the transfer of information between feature fusion and GRU~\cite{cho2014learning} modules continue to hinder performance.

\begin{figure*}[!h] 
    \centering 
    \centering 
    \includegraphics[width=\textwidth]{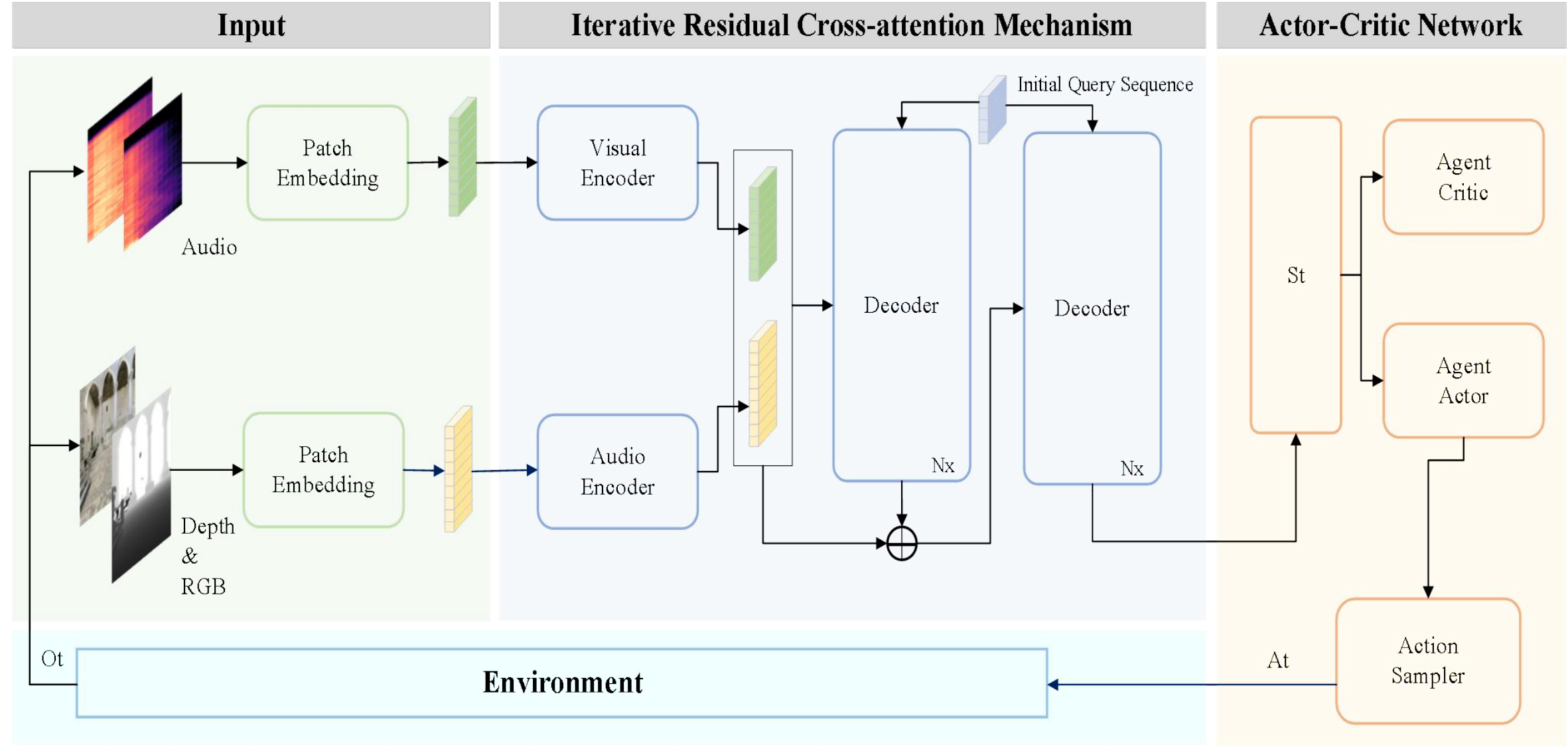} 
    \caption{Audio-Visual Navigation Network. Our model leverages acoustic and visual cues from the 3D environment to achieve efficient navigation through a novel iterative residual cross-attention mechanism.}
    \label{fig2}
\end{figure*}

To address the existing challenges in audiovisual navigation, we introduce the IRCAM-AVN (Iterative Residual Cross-Attention Mechanism for Audiovisual Navigation) model, designed to establish an integrated and efficient framework for this purpose. As illustrated in Fig.~\ref{fig1}, our integrated module synthesizes the functionalities of traditional feature fusion and GRU~\cite{cho2014learning} (Gated Recurrent Unit) modules. To enhance success rates and the timeliness of audiovisual navigation tasks, our model innovatively employs multiple residual designs to fully leverage multimodal information for extracting navigation cues. In summary, our principal contributions are as follows:

\begin{itemize} 
\item We propose an integrated framework (IRCAM-AVN) that substitutes the traditional feature fusion and GRU~\cite{cho2014learning} modules utilized in audiovisual navigation tasks with the Iterative Residual Cross-Attention Module. This advancement innovatively combines feature fusion and sequence modeling into a single comprehensive module. 

\item We introduce a novel method for attention computation. Our multiple residual module facilitates multi-level information interaction during cross-modal processing. In addition to incorporating residual structures within both the encoder and decoder, we creatively establish external residual connections among encoders to mitigate significant prediction bias throughout the inference process effectively.

\item We perform empirical evaluations on two real-world 3D environment datasets, Replica~\cite{straub2019replica} and Matterport3D~\cite{chang2017matterport3d}, thereby validating the effectiveness and robustness of our intelligent agent in locating audio targets successfully. 

\end{itemize}

\section{RELATED WORK}

Research in embodied navigation~\cite{chaplot2020neural,mezghan2022memory,chen2025affordances,zhang2025mapnav} primarily aims to equip intelligent agents with the capability to navigate intricate environments utilizing a combination of sensory modalities. Traditional methodologies~\cite{chaplot2020neural,shah2021ving} have predominantly relied on visual inputs; however, recent advancements~\cite{chen2020soundspaces,chen2021waypoints} have successfully integrated auditory signals, facilitating multimodal information fusion and a more comprehensive capture of contextual information. Notably, recent studies~\cite{chen2023omnidirectional, YinfengIJCAI2023MACMA,ttt, SAAVN} have introduced audio-rendered three-dimensional simulation environments tailored for audio-visual navigation tasks. These innovations highlight the critical need for effective processing of multimodal information to enhance information fusion and decision-making processes. Nevertheless, existing advanced techniques often require sequential modeling to adequately capture temporal dependencies after implementing fusion strategies. In contrast, our proposed approach introduces architectural enhancements to current audio-visual navigation tasks.

The Transformer architecture~\cite{vaswani2017attention} employs a multi-head self-attention mechanism, which retains contextual interactions throughout the entire sequence and allows for parallel processing of inputs, thereby directly modeling long-range dependencies. This design significantly accelerates training and inference in scenarios involving large-scale data and deep neural networks. Conversely, the Gated Recurrent Unit~\cite{cho2014learning} updates its hidden state through a sequential, time-step-by-time-step process, a dependency that restricts parallelization and makes it susceptible to challenges such as vanishing gradients and information decay when modeling long sequences. 

\section{APPROACH}

In this study, we propose the IRCAM (Iterative Residual Cross-Attention Module) for audiovisual navigation. As illustrated in Fig.~\ref{fig2}, our model departs from conventional audiovisual navigation approaches by primarily incorporating two fundamental components. First, multimodal information is processed through a patch embedding layer, generating embedding sequences. These sequences subsequently undergo a self-attention mechanism, which assigns attention weights to various features, resulting in the formation of the Initial Multimodal Sequence. Following processing through the decoder, the output is concatenated with the initial sequence and serves as the input for the subsequent decoding step. Second, we employ an actor-critic network to facilitate action prediction, evaluation, and optimization. The robotic agent iteratively executes this process until it successfully acquires the moving audio target. We will elaborate on the core components in the forthcoming sections.

At each time step \( t \), the agent receives multimodal observations:

\begin{equation}
O_t = (I_t, B_t),
\end{equation}

where \( I_t \) and \( B_t \) represent visual and audio inputs, respectively. These inputs are embedded into a unified multimodal sequence via patch embedding:

\begin{equation}
E_t = \text{Concat}(\text{Audio Encoder}(I_t), \text{Visual Encoder}(B_t)),
\end{equation}

The sequence $E_t$ is used as the key-value memory input to the decoder, and the information extraction is performed under the guidance of the initial query sequence $Q$. Here is the core formula:

\begin{equation}
\text{MultiHeadAttention}(Q, E_tK, E_tV) = \text{Concat}(\text{head}_i) W^O,
\end{equation}

\begin{equation}
\text{head}_i = \text{Attention}(Q W^Q_i, E_tK W^K_i, E_tV W^V_i),
\end{equation}

\begin{equation}
\text{Attention}(Q, E_tK, E_tV) = \text{softmax}\left( \frac{Q E_tK^T}{\sqrt{d_k}} \right) E_tV.
\end{equation}

After decoding, the updated sequence is constructed as:

\begin{equation}
E_{t+1} = \text{Concat}(\text{Decoder}(E_t), E_t).
\end{equation}

The integrated sequence \( E_{t+1} \) serves as the input for subsequent decoder operations. This multi-level residual structure effectively reduces bias and ensures efficient transmission of information.


Why Iterative Residual Cross-Attention can replace Feature Fusion and GRU~\cite{cho2014learning}?

Limitations of Traditional Methods. Traditional audiovisual navigation approaches often employ convolutional operations for feature fusion and GRU~\cite{cho2014learning} (Gated Recurrent Unit) modules for temporal modeling in separate stages. However, convolutional operations have limited capacity for extracting global features. Although increasing the number of convolutional layers or using dilated convolutions can enlarge the receptive field, these methods raise computational costs and risk issues like vanishing gradients. Moreover, convolution is inherently less suited for dynamic and effective feature fusion.

Issues in State-of-the-Art Models. Recent state-of-the-art models commonly adopt attention mechanisms~\cite{Yu_2022_BMVC} or their variants for early-stage information fusion, followed by GRU~\cite{cho2014learning} modules for sequence modeling. This two-stage design can result in redundant modeling of long-range dependencies, as part of the sequence modeling is already implicitly performed during the early fusion phase.

Advantages of the Proposed Method. The proposed Iterative Residual Cross-Attention Mechanism not only achieves dynamic feature fusion and enhancement but also addresses the problem of functional redundancy. By creatively introducing a multi-level residual structure, the model autonomously corrects deviations and dynamically adjusts association weights among features, enabling more effective multimodal fusion and capturing of long-range dependencies compared to previous approaches.

\begin{table*}[htbp]
\centering

\caption{Comparison of audio-visual navigation results on the Replica and Matterport3D datasets.}
\begin{tabular}{@{}l|cccccc|cccccc@{}}
\toprule
                                         & \multicolumn{6}{c|}{Replica}                                                                                           & \multicolumn{6}{c}{Matterport3D}                                                                                       \\ \cmidrule(l){2-13} 
\multicolumn{1}{c|}{Method}              & \multicolumn{3}{c|}{Heard}                                           & \multicolumn{3}{c|}{Unheard}                    & \multicolumn{3}{c|}{Heard}                                           & \multicolumn{3}{c}{Unheard}                     \\ \cmidrule(l){2-13} 
                                         & SNA $\uparrow$ & SR $\uparrow$ & \multicolumn{1}{c|}{SPL $\uparrow$} & SNA $\uparrow$ & SR $\uparrow$ & SPL $\uparrow$ & SNA $\uparrow$ & SR $\uparrow$ & \multicolumn{1}{c|}{SPL $\uparrow$} & SNA $\uparrow$ & SR $\uparrow$ & SPL $\uparrow$ \\ \midrule
Random Agent~\cite{chen2021waypoints}         & 1.8            & 18.5          & \multicolumn{1}{c|}{4.9}            & 1.8            & 18.5          & 4.9            & 0.8            & 9.1           & \multicolumn{1}{c|}{2.1}            & 0.8            & 9.1           & 2.1            \\
Direction Follower~\cite{chen2021waypoints}    & 41.1           & 72.0          & \multicolumn{1}{c|}{54.7}           & 8.4            & 17.2          & 11.1           & 23.8           & 41.2          & \multicolumn{1}{c|}{32.3}           & 10.7           & 18.0          & 13.9           \\
Frontier Waypoints~\cite{chen2021waypoints}    & 35.2           & 63.9          & \multicolumn{1}{c|}{44.0}           & 5.1            & 14.8          & 6.5            & 22.2           & 42.8          & \multicolumn{1}{c|}{30.6}           & 8.1            & 16.4          & 10.9           \\
Supervised Waypoints~\cite{chen2021waypoints}  & 48.5           & 88.1          & \multicolumn{1}{c|}{59.1}           & 10.1           & 43.1          & 14.1           & 16.2           & 36.2          & \multicolumn{1}{c|}{21.0}           & 2.9            & 8.8           & 4.1            \\
Gan et al.~\cite{gan2020look}           & 47.9           & 83.1          & \multicolumn{1}{c|}{57.6}           & 5.7            & 15.7          & 7.5            & 17.1           & 37.9          & \multicolumn{1}{c|}{22.8}           & 3.6            & 10.2          & 5.0            \\
AV-Nav~\cite{chen2020soundspaces}              & 52.7           & 94.5          & \multicolumn{1}{c|}{78.2}           & 16.7           & 50.9          & 34.7           & 32.6           & 71.3          & \multicolumn{1}{c|}{55.1}           & 12.8           & 40.1          & 25.9           \\
AV-WaN~\cite{chen2021waypoints}                & 70.7           & \textbf{98.7}          & \multicolumn{1}{c|}{86.6}           & 27.1           & 52.8          & 34.7           & 54.8           & 93.6          & \multicolumn{1}{c|}{72.3}           & 30.6           & 56.7          & 40.9           \\
ORAN~\cite{chen2023omnidirectional}                 & 70.1           & 96.7          & \multicolumn{1}{c|}{84.2}           & 36.5           & 60.9          & 46.7           & 57.7           & 93.5          & \multicolumn{1}{c|}{73.7}           & 35.3           & 59.4          & 50.8           \\
IRCAM (ours)                             & \textbf{73.2}  & 97.6 & \multicolumn{1}{c|}{\textbf{89.9}}  & \textbf{41.6}  & \textbf{64.4} & \textbf{51.9}  & \textbf{61.4}  & \textbf{95.3} & \multicolumn{1}{c|}{\textbf{80.7}}  & \textbf{37.8}  & \textbf{61.5} & \textbf{52.4}  \\ \bottomrule
\end{tabular}
\label{tab:tab1}
\end{table*}

\section{EXPERIMENT}

\subsection{EXPERIMENTAL SETUP}
In this section, we describe the experimental setup used to evaluate the performance of our proposed IRCAM architecture for achieving faster audiovisual embodied navigation in 3D environments. Our experiments are designed to demonstrate the effectiveness of our method in integrating audio and visual modalities, which is based on a multi-residual cross-attention mechanism and an early-fusion approach.

Dataset. We evaluated our proposed model on the Replica~\cite{straub2019replica} and Matterport3D~\cite{chang2017matterport3d} datasets. The Replica dataset~\cite{straub2019replica} contains 18 high-resolution 3D scenes captured using professional scanning equipment, providing RGB images, depth maps, semantic annotations, camera poses, and high-fidelity 3D meshes. The Matterport3D dataset~\cite{chang2017matterport3d} is a large-scale collection of real indoor environments, comprising 90 building-scale scenes with an average floor area of approximately 517 square meters. Following SoundSpaces, we select a subset of 85 scenes for our audio-rendering experiments.

Experimental Environment. Our experimental setup builds on the pioneering SoundSpaces framework, which integrates the open-source Habitat~\cite{savva2019habitat} platform with the Replica~\cite{straub2019replica} and Matterport3D~\cite{chang2017matterport3d} datasets. By incorporating room impulse responses (RIRs), bidirectional path-tracing algorithms, and material property configurations for audio simulation, we construct a fully audio-rendered 3D environment in which agents can learn navigation policies and accurately navigate to target sound sources in novel, unlabeled scenes.

Implementation Details. For the experiments on the Replica dataset~\cite{straub2019replica}, we set the learning rate to \(1 \times 10^ {-4}\). However, for the Matterport3D dataset~\cite{chang2017matterport3d}, we found it necessary to use a learning rate of \(4 \times 10^ {-5}\). We optimize the model with the Adam optimizer~\cite{KingmaB14}, using PyTorch~\cite{PaszkeGMLBCKLGA19}’s default momentum coefficients. Each training episode is limited to a maximum of 500 actions. We apply a reward discount factor of 0.99. We trained the model for 80,000 agent steps on the Replica dataset~\cite{straub2019replica} and 100,000 agent steps on the Matterport3D dataset~\cite{chang2017matterport3d}.

Evaluation Metrics. We use Success Rate (SR), Success Weighted by Inverse Path Length (SPL), and Success Weighted by Inverse Number of Actions (SNA) as evaluation metrics.


\subsection{Performance Comparison}
In this work, we propose a novel and practical framework for audiovisual embodied navigation in 3D environments. To demonstrate the effectiveness of the proposed IRCAM, we compare it against previous audio-visual embodied navigation baselines~\cite{chen2020soundspaces,chen2023omnidirectional,chen2022soundspaces,gan2020look}.

For the Replica~\cite{straub2019replica} dataset, we present quantitative comparison results in Table~\ref{tab:tab1}. As shown, our method achieves the best performance on all metrics in both “Heard” and “Unheard” settings compared to previous audio-visual navigation approaches. In particular, our IRCAM outperforms ORAN~\cite{chen2023omnidirectional}—which represents the current state-of-the-art audio-visual navigation baseline—by 3.1 SNA@Heard, 0.9 SR@Heard, and 5.7 SPL@Heard, as well as by 5.1 SNA@Unheard, 3.5 SR@Unheard, and 5.2 SPL@Unheard across the two settings. Moreover, compared to AV-WaN~\cite{chen2021waypoints}, the leading waypoint-based baseline, we achieve substantial gains, highlighting the importance of our unified IRCAM module for the audio-visual navigation task.

Additionally, Table~\ref{tab:tab1} shows significant improvements on the Matterport3D~\cite{chang2017matterport3d} benchmark. Against AV-WaN~\cite{chen2021waypoints}, the current state-of-the-art waypoint-based method, we achieve gains of 6.6 SNA@Heard, 1.7 SR@Heard, and 8.4 SPL@Heard, as well as 7.2 SNA@Unheard, 4.8 SR@Unheard, and 11.5 SPL@Unheard in the two settings. These results demonstrate that our model effectively captures multimodal fusion and temporal information for audio-visual navigation.

\begin{figure}[!h] 
    \centering 
    \vspace{0.001cm} 
    \centering 
    \includegraphics[width=0.5\textwidth]{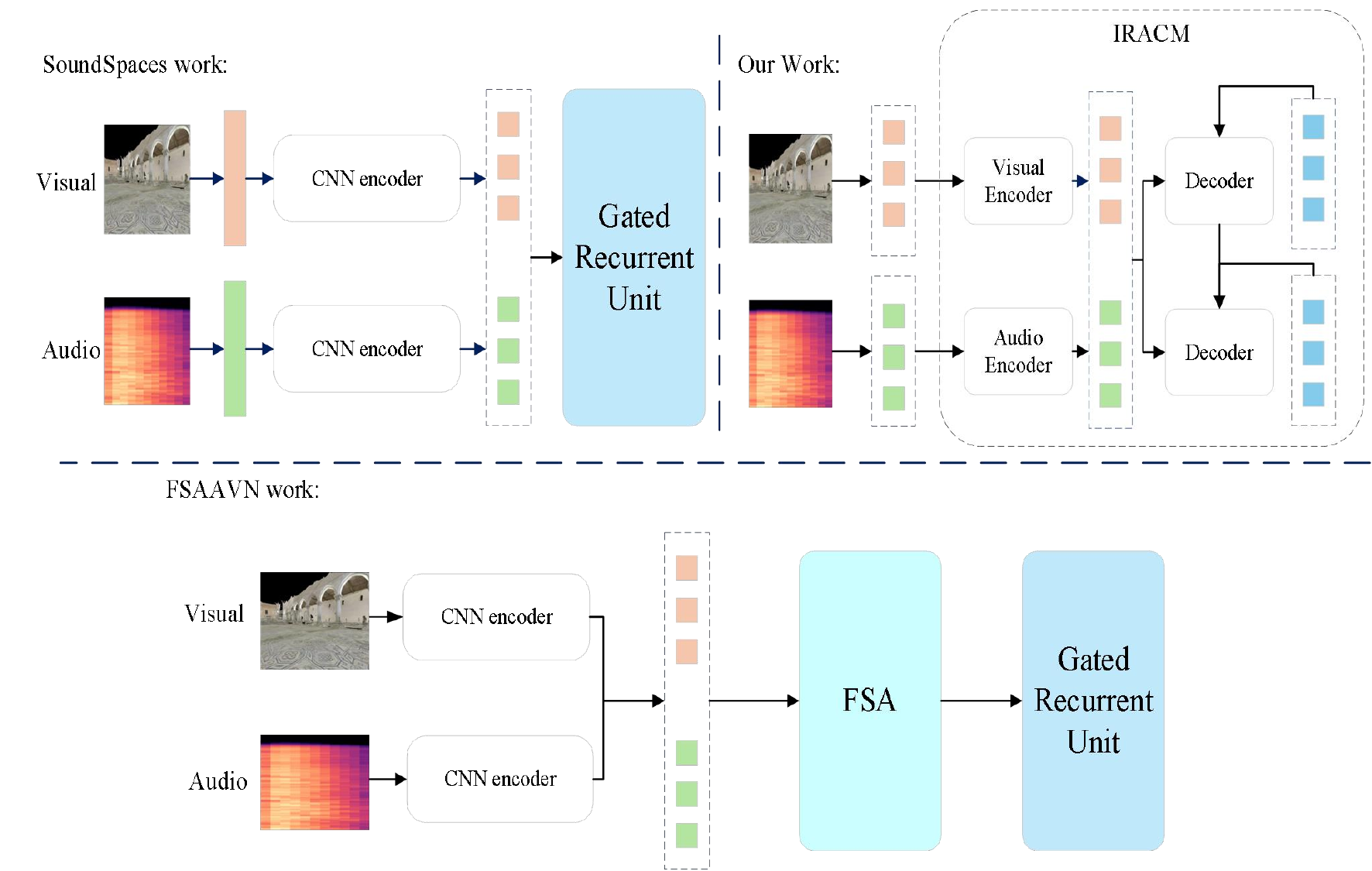} 
    \caption{Our audiovisual embodied navigation architecture is compared with previous works on embodied navigation in 3D environments. Earlier audiovisual navigation methods typically perform feature fusion and GRU operations in separate stages. However, this staged processing often results in uncoordinated information flow and insufficient feature integration, ultimately leading to inefficient pathfinding.}
    \label{fig3}
\end{figure}

\begin{figure}[h] 
    \centering 
    \includegraphics[width=0.5\textwidth]{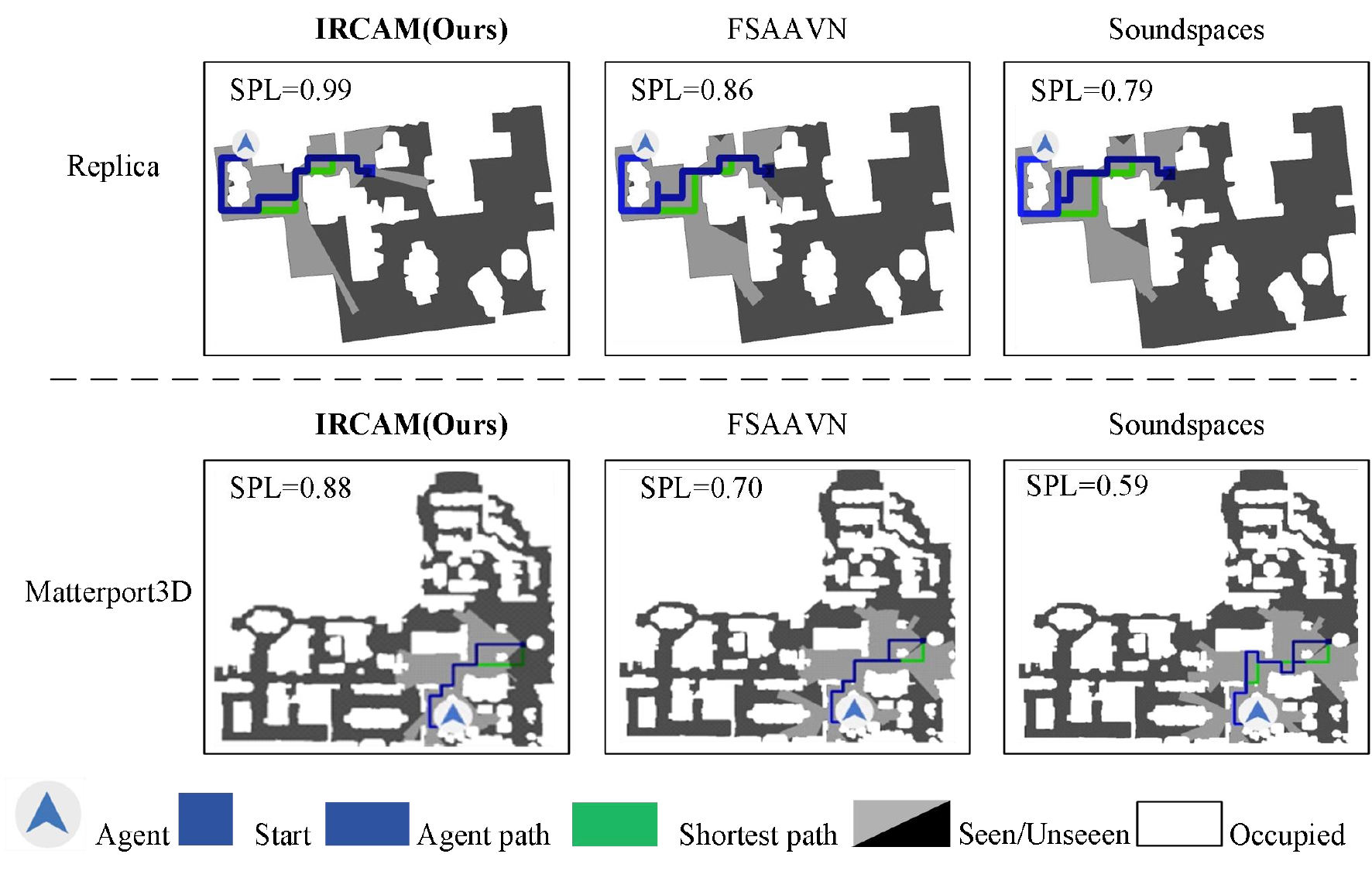}
    \caption{An intuitive comparison between our model and various baselines on the audio-visual navigation task in the Replica and Matterport3D datasets.}
    \label{fig4}
\end{figure}

\begin{table}[]
\centering 
\caption{Comparison of Different Methods on the Replica and Matterport3D datasets.}
\label{tab:tab2}
\resizebox{0.5\textwidth}{!}{%
\begin{tabular}{@{}l|l|cc|cc@{}}
\toprule
\textbf{}                   & \textbf{}                    & \multicolumn{2}{c|}{Replica}      & \multicolumn{2}{c}{Matterport3D}  \\ \cmidrule(l){3-6} 
\multicolumn{1}{c|}{Models} & \multicolumn{1}{c|}{Methods} & Heard           & Unheard         & Heard           & Unheard         \\
                            &                              & SPL($\uparrow$) & SPL($\uparrow$) & SPL($\uparrow$) & SPL($\uparrow$) \\ \midrule
SoundSpaces                 & CNN+GRU                      & 78.2            & 34.7            & 55.1            & 25.9            \\
FSAAVN                      & SelfAttention+GRU            & 80.9            & 42.3            & 68.2            & 33.5            \\
IRACM(Ours)                 & IRACM                        & \textbf{89.9}   & \textbf{51.9}   & \textbf{80.7}   & \textbf{52.4}   \\ \bottomrule
\end{tabular}
}
\end{table}

\subsection{Comparison of Different Methods}
 
As shown in Fig.~\ref{fig3}, our proposed IRCAM module outperforms existing audiovisual navigation methods in predicting navigation trajectories. As illustrated in the architecture diagram in Fig.~\ref{fig4}, traditional audiovisual navigation frameworks typically consist of a feature fusion module and a GRU-based sequence modeling module. In contrast, our model introduces a novel IRCAM module that innovatively integrates feature fusion and sequential modeling into a unified structure. Specifically, as reported in Table~\ref{tab:tab2}, the IRCAM module significantly outperforms FSAAVN~\cite{Yu_2022_BMVC} and SoundSpaces~\cite{chen2020soundspaces}, both of which adopt a staged approach to audiovisual navigation, across all evaluation metrics.

\subsection{Ablation Study}
We conducted ablation studies on the Replica~\cite{straub2019replica} and Matterport3D~\cite{chang2017matterport3d} datasets. Detailed performance comparisons are presented in Table~\ref{tab:tab3} and Table~\ref{tab:tab4}.

\subsubsection{w/o RT}
The RT module is removed, meaning that the information sequence after the decoder is no longer concatenated with the Initial Multimodal Sequence, and subsequent decoder operations do not process it. Without the RT module, the model cannot perform incremental refinements based on previously generated information, which prevents it from achieving a deep understanding and error correction.

\subsubsection{w/o PE}
The PE module is removed, meaning that audio information is no longer processed through Patch Embedding but instead through convolutional operations. Without the PE module, the model cannot achieve structural and semantic alignment, thus failing to provide properly preprocessed data for subsequent modules.

\begin{figure}[!h] 
    \centering 
    \vspace{0.05cm} 
    \centering 
    \includegraphics[width=0.5\textwidth]{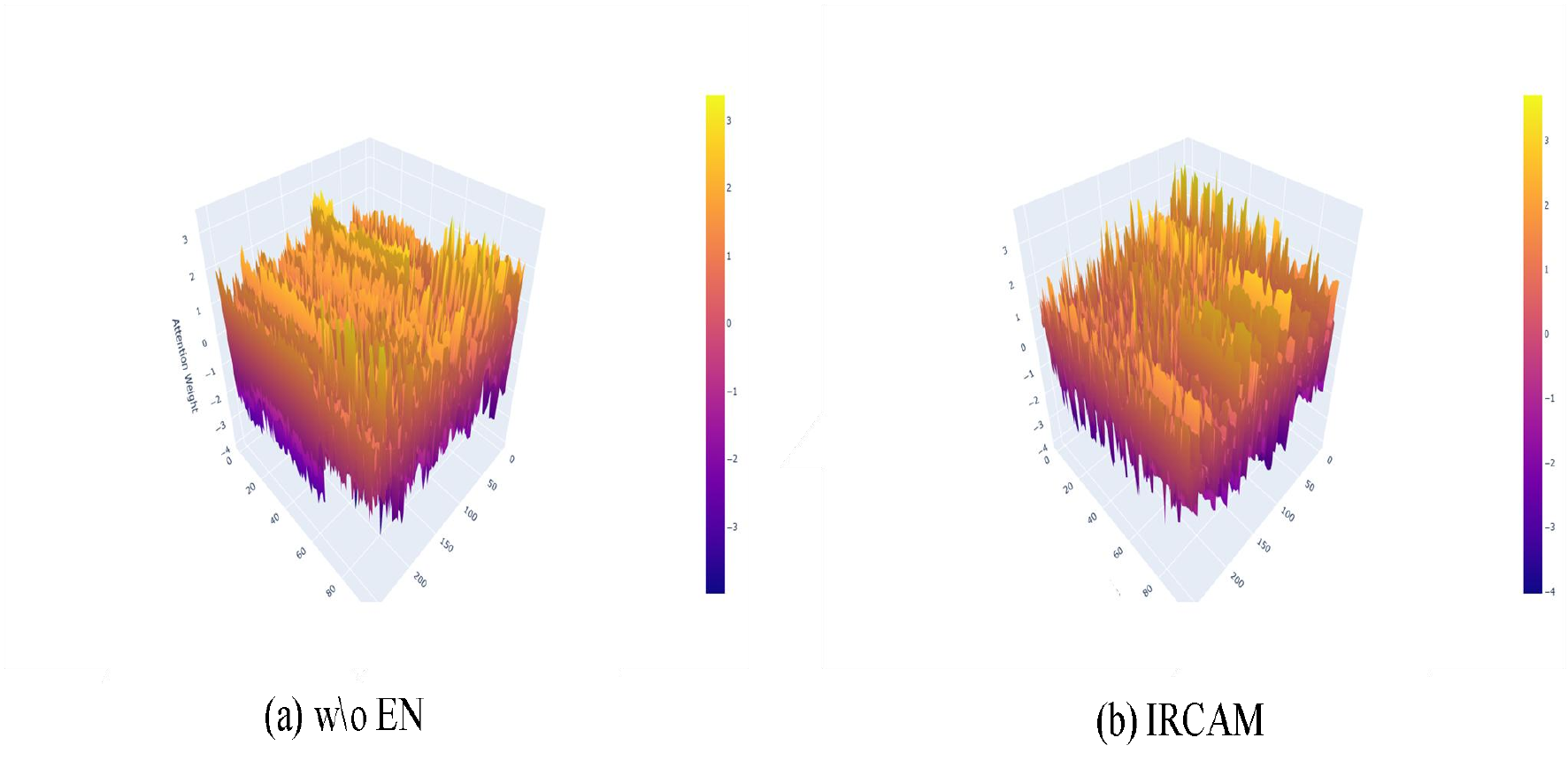} 
    \caption{Our audiovisual embodied navigation architecture is also compared with a version of the architecture where the EN module is ablated. It is observed that without the EN module, the attention mechanism struggles to distinguish and focus on critical regions effectively.}
    \label{fig5}
\end{figure}

\subsubsection{w/o EN}
The EN module is removed, meaning that the multimodal information sequence bypasses the Encoder and directly uses the embedded sequence after Patch Embedding as the Initial Multimodal Sequence for subsequent operations. As shown in Fig.~\ref{fig5}, without the EN module, the model is unable to capture key information within the multimodal inputs, resulting in insufficient deep preprocessing of the multimodal information sequence.

These ablation results demonstrate that each module is indispensable and that all of them contribute significantly to the model’s performance on the audiovisual navigation task.

\begin{table}[]
\caption{Ablation study on the proposed model on Replica.}
\label{tab:tab3}
\resizebox{0.5\textwidth}{!}{%
\begin{tabular}{@{}c|cccccc@{}}
\toprule
            & \multicolumn{6}{c}{Replica}                                                                                                  \\ \cmidrule(l){2-7} 
Models      & \multicolumn{3}{c|}{Heard}                                              & \multicolumn{3}{c}{Unheard}                        \\ \cmidrule(l){2-7} 
\textbf{}   & SNA($\uparrow$) & SR($\uparrow$) & \multicolumn{1}{c|}{SPL($\uparrow$)} & SNA($\uparrow$) & SR($\uparrow$) & SPL($\uparrow$) \\ \midrule
w/o EN      & 38.3            & 84.1           & \multicolumn{1}{c|}{67.2}            & 11.2            & 44.9           & 23.4            \\
w/o PE      & 52.9            & 94.1           & \multicolumn{1}{c|}{79.3}            & 24.7            & 53.5           & 35.6            \\
w/o RT      & 70.6            & 96.3           & \multicolumn{1}{c|}{84.4}            & 36.5            & 59.8           & 46.1            \\
IRCAM(Ours) & \textbf{73.2}   & \textbf{97.6}  & \multicolumn{1}{c|}{\textbf{89.9}}   & \textbf{41.6}   & \textbf{64.4}  & \textbf{51.9}   \\ \bottomrule
\end{tabular}
}
\par\vspace{0.5em}
\textbf{Note:} “w/o” denotes without.
\end{table}

\begin{table}[]
\caption{Ablation study on the proposed model on Matterport3D}
\label{tab:tab4}
\resizebox{0.5\textwidth}{!}{%
\begin{tabular}{@{}c|cccccc@{}}
\toprule
            & \multicolumn{6}{c}{Matterport3D}                                                                                             \\ \cmidrule(l){2-7} 
Models      & \multicolumn{3}{c|}{Heard}                                              & \multicolumn{3}{c}{Unheard}                        \\ \cmidrule(l){2-7} 
\textbf{}   & SNA($\uparrow$) & SR($\uparrow$) & \multicolumn{1}{c|}{SPL($\uparrow$)} & SNA($\uparrow$) & SR($\uparrow$) & SPL($\uparrow$) \\ \midrule
w/o EN      & 24.7            & 66.6           & \multicolumn{1}{c|}{47.9}            & 11.4            & 36.6           & 21.5            \\
w/o PE      & 52.3            & 91.4           & \multicolumn{1}{c|}{69.8}            & 23.5            & 53.2           & 31.6            \\
w/o RT      & 58.1            & 93.7           & \multicolumn{1}{c|}{74.4}            & 34.1            & 58.3           & 49.5            \\
IRCAM(Ours) & \textbf{61.4}   & \textbf{95.3}  & \multicolumn{1}{c|}{\textbf{80.7}}   & \textbf{37.8}   & \textbf{61.5}  & \textbf{52.4}   \\ \bottomrule
\end{tabular}
}
\par\vspace{0.5em}
{\textbf{Note:} “w/o” denotes without.} 
\end{table}

\subsection{Visualization}

\begin{figure}[!h] 
    \centering 
    \vspace{0.05cm} 
    \centering 
    \includegraphics[width=0.5\textwidth]{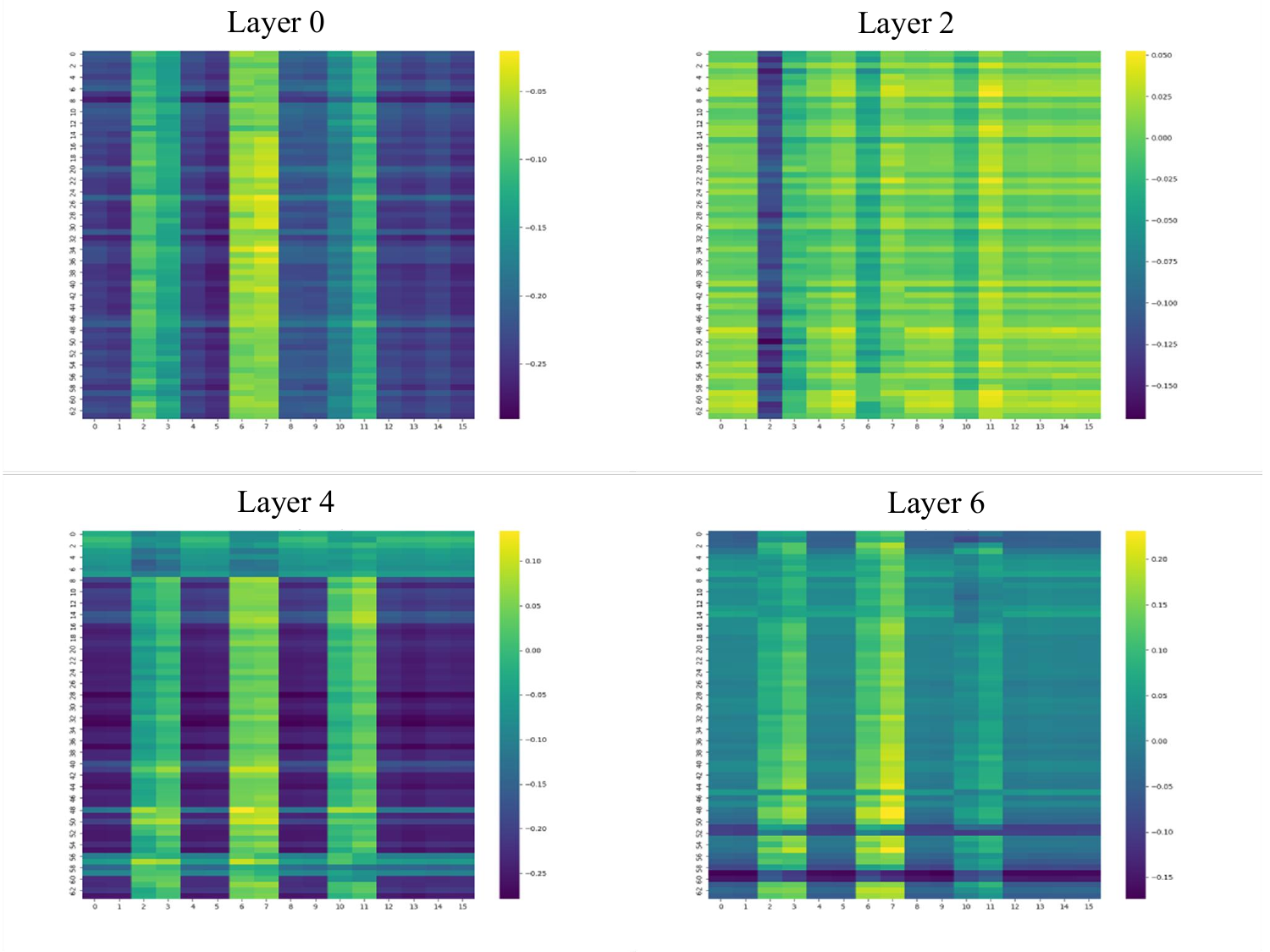} 
    \caption{The number of decoder layers has a significant impact on the correlation between visual and audio feature vectors. Experimental results show that as the number of decoder layers increases, the correlation between the two gradually strengthens, with the correlation range expanding and tending toward higher positive values.}
    \label{fig6}
\end{figure}

In the IRCAM-AVN framework, the visual encoder generates a sequence of visual feature vectors, while the auditory encoder produces corresponding audio feature vectors. This forms the basis for multimodal perception. Throughout the iterative decoding process, information from the two modalities is gradually integrated and refined, ultimately resulting in feature vectors characterized by enhanced clarity in both structure and meaning. To effectively illustrate this process, attention heatmaps are employed to analyze the correlation between feature vectors from different modalities, which are subsequently projected into a two-dimensional space for visualization purposes. A comparative analysis of feature distributions across various iterations and decoder depths reveals that the range of modality correlation is maximized when using a six-layer decoder. At this depth, the information structure becomes markedly more concentrated and well-defined. Fig.~\ref{fig6} presents the experimental findings, showcasing the dynamic evolution of multimodal features throughout different stages of decoding. As the depth of decoding increases, the correlation between visual and auditory features intensifies significantly, leading to a more compact feature distribution rather than a dispersed one. This indicates that the model achieves proficient multimodal feature fusion at deeper decoding stages, thereby providing more discriminative feature representations for subsequent tasks.

\section{CONCLUSIONS}

In this study, we present IRCAM, an innovative method designed to enhance the efficiency of audiovisual embodied navigation within three-dimensional environments. This approach substitutes traditional feature-fusion and GRU modules with a unified IRCAM block, facilitating effective feature integration while also capturing long-range sequence dependencies. Through its multi-residual and cross-attention mechanisms, our model adeptly identifies the critical information necessary for effective audiovisual navigation. Experimental findings conducted on the Replica and Matterport3D datasets indicate that our method surpasses various state-of-the-art baselines, demonstrating that advanced multimodal fusion strategies can yield more robust and adaptable navigation solutions.

\addtolength{\textheight}{-12cm}   




\section*{ACKNOWLEDGMENT}

This research was financially supported by the National Natural Science Foundation of China (Grants Nos. 62463029, 62472368, and 62303259) and the Natural Science Foundation of Tianjin (Grant No. 24JCQNJC00910).


\bibliographystyle{IEEEtran}   
\bibliography{mybibfile}      

\end{document}